%% file: main.tex
\documentclass[10pt,twocolumn,letterpaper]{article}

\usepackage{cvpr}              %

\usepackage[pagebackref,breaklinks,colorlinks,allcolors=cvprblue]{hyperref}

\usepackage{url}
\usepackage[utf8]{inputenc} %
\usepackage[T1]{fontenc}    %
\usepackage{hyperref}       %
\usepackage{url}    %
\usepackage{booktabs}       %
\usepackage{amsfonts}       %
\usepackage{nicefrac}       %
\usepackage{microtype}      %
\usepackage{xcolor} %

\usepackage{graphicx}
\usepackage{booktabs}
\usepackage{tabularx}
\usepackage{epsfig}
\usepackage{graphicx}
\usepackage{amsmath}
\usepackage{bm}
\usepackage{amssymb}
\usepackage{colortbl}
\usepackage{multirow}
\usepackage{color}
\usepackage{setspace}
\usepackage[normalem]{ulem}
\usepackage{tikz}
\usepackage{comment}
\usepackage{algorithm}
\usepackage{algpseudocode}
\usepackage{float}
\usepackage{caption}
\usepackage{adjustbox}
\usepackage{pifont}
\usepackage{subcaption}
\usepackage{rotating}
\usepackage{makecell}
\usepackage[accsupp]{axessibility}  %
\usepackage{wrapfig}

\definecolor{top1}{RGB}{245,152,153}
\definecolor{top2}{RGB}{253,205,154}
\definecolor{top3}{RGB}{248,244,140}

\usepackage[accsupp]{axessibility}  %

\definecolor{blgray}{gray}{0.97}
\definecolor{mygray}{gray}{.93}

\usepackage[capitalize]{cleveref}
\Crefname{section}{Section}{Sections}
\Crefname{table}{Table}{Tables}

\input{preamble}
\definecolor{cvprblue}{rgb}{0.21,0.49,0.74}

\definecolor{nicegreen}{rgb}{0.1, 0.6, 0.2}

\def\paperID{13541} %
\def\confName{CVPR}
\def\confYear{2025}

\def\ourmodel{VisAH\xspace} 
\def\ourdataset{\textsc{the muddy mix dataset}\xspace}

\title{Learning to Highlight Audio by Watching Movies}

\author{
Chao Huang{$^1$}, Ruohan Gao{$^2$}, J. M. F. Tsang{$^3$}, Jan Kurcius{$^3$}, Cagdas Bilen{$^3$}, \\Chenliang Xu{$^1$}, Anurag Kumar{$^3$}, Sanjeel Parekh{$^3$} \\
{$^1$}University of Rochester, {$^2$} University of Maryland, College Park, {$^3$}Meta Reality Labs Research\\
}

\begin{document}

\maketitle
\input{sec/0_abstract}

\input{sec/1_intro}

\input{sec/2_background}
\input{sec/3_method}
\input{sec/4_dataset}

\input{sec/5_experiments}
\input{sec/6_conclusion}

{
    \small
    \bibliographystyle{ieeenat_fullname}
    \bibliography{main}
}
\input{sec/X_suppl}

\end{document}

%% file: sec/0_abstract.tex
\begin{abstract}

Recent years have seen a significant increase in video content creation and consumption.
Crafting engaging content requires the careful curation of both visual and audio elements. While visual cue curation, through techniques like optimal viewpoint selection or post-editing, has been central to media production, 
its natural counterpart, audio, has not undergone equivalent advancements. This often results in a disconnect between visual and acoustic saliency. To bridge this gap, we introduce a novel task: visually-guided acoustic highlighting, which aims to transform audio to deliver appropriate highlighting effects guided by the accompanying video, ultimately creating a more harmonious audio-visual experience. We propose a flexible, transformer-based multimodal framework to solve this task. To train our model, we also introduce a new dataset---\ourdataset, leveraging the meticulous audio and video crafting found in movies,
 which provides a form of free supervision. We develop a pseudo-data generation process to simulate poorly mixed audio, mimicking real-world scenarios through a three-step process---separation, adjustment, and remixing. Our approach consistently outperforms several baselines in both quantitative and subjective evaluation. We also systematically study the impact of different types of contextual guidance and difficulty levels of the dataset. Our project page is here: \url{https://wikichao.github.io/VisAH/}.

\end{abstract}

%% file: sec/1_intro.tex
\section{Introduction}
\label{sec:intro}

\begin{figure}[!ht]
    \centering
    \includegraphics[width=1\linewidth]{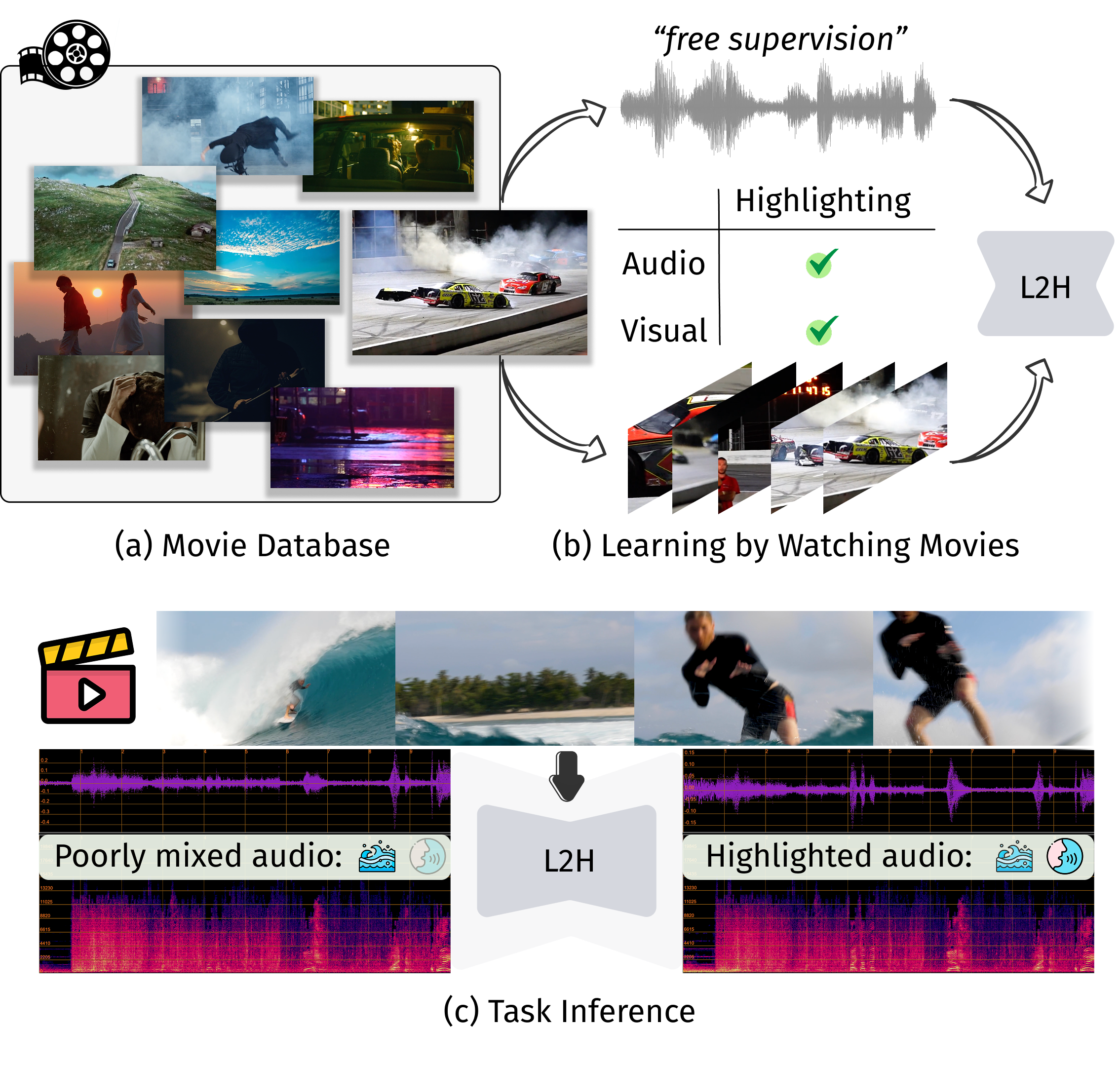}
    \caption{We propose a new task that aims to transform poorly mixed audio into a well-balanced mix using visual guidance. One of our key insights is to use well-curated audio-visual content from a movie database as free supervision to learn the appropriate highlighting effect for audio (L2H).  %
    }
    \label{fig:teaser}
    \vspace{-0.1in}
\end{figure}

Be it amateur recordings of memorable moments or professionally created content---telling a story and delivering the best audio-visual experience with a video requires the right balance of audio and visual elements in the scene. Consider for example, the scene in Fig.~\ref{fig:teaser}c that depicts a video of a man talking in the sea. The scene is best represented when the focus is on the person and the sea at the appropriate moments. While there are several ways to visually highlight the intended objects during capture or in post-production \cite{rao2023dynamic, wang2024lave}, they remain relatively under-explored and limited for the acoustics. In our case above, the man's speech can be obscured by the sound of crashing waves. Is it possible to automatically adjust the levels of the speech and the wave sounds according to the video content to ensure they are both acoustically salient and well-balanced?

A naive approach would involve first demixing the sounds into different source components and then remixing them at their respective intended levels. But doing so has two major drawbacks: (i) imperfect demixing could lead to the highlighting of undesired sources, and (ii) ensuring the right  temporal variations and alignment with the video manually is a laborious process.
Some research efforts, such as in music remixing~\cite{yang2022don, lu2019play,koo2023music}, focus on adjusting the levels and effects of individual instruments to recompose music tracks. However, this approach is limited to the music domain, neglecting the broader needs of natural audio composition across varied media contexts.  

In this paper, we aim to bridge this gap by introducing a novel task, \textbf{visually-guided acoustic highlighting}. Our approach builds on the hypothesis that the visual stream in media is often curated with intent, implicitly conveying highlighted content. In contrast, due to the limitations of recording devices, such as microphones attached to video cameras that capture all sounds indiscriminately, audio often lacks intentional mixing, resulting in a poorly balanced track. The goal of our task is to use the video as guidance to transform the poorly mixed audio with appropriate highlighting effects, ensuring a better output audio mix.

Dataset is a key requirement for training a learning-based model to perform this task. We observe that movies are inherently well-curated where audio is meticulously crafted alongside video to create intentional highlighting effects. This provides us with free supervision for the acoustic highlighting effect as illustrated in Fig.~\ref{fig:teaser}. Consequently, we build a new dataset using movie clips spanning several generes from the Condensed Movie Dataset (CMD)~\cite{bain2020condensed}. To simulate real-world scenarios where audio may be poorly mixed or require enhancement, we introduce a pseudo-data generation process that begins with high-quality movie audio and then applies imperfect separation, followed by adjustment and remixing of individual audio sources.

 We tackle acoustic highlighting as an audio-to-audio translation problem and propose a transformer-based Visually-guided Acoustic Highlighting (\ourmodel) model. \ourmodel uses a U-Net-like audio backbone with a dual encoder~\cite{defossez2021hybrid} that takes both the spectrogram and the waveform as inputs to extract latent representations from the poorly mixed audio. In this latent space, a transformer encoder processes context, such as the video stream or its corresponding caption, and a transformer decoder with cross-attention integrates this video context to guide the transformation. The decoder then converts the poorly mixed audio representation into highlighted audio. This design is flexible, supporting easy adjustments in both the backbone and latent modules. Specifically, our model combines video context encoding and visually-guided audio decoding, enabling it to fully capture temporal and semantic trends in the video and leverage them for effective audio highlighting. 

To summarize, our main contributions are threefold:
\begin{itemize} 
\item We propose to intelligently highlight the audio content in a video guided by visual cues, and we design  \ourmodel, a mutimodal transformer-based model to achieve that goal.
\item Leveraging the free supervision from movies, where both audio and video are already meticulously crafted, we introduce \ourdataset, a new dataset curated for this task.
\item Our method outperforms a series of baselines, effectively highlighting audio across different types of video content.
\end{itemize}

%% file: sec/2_background.tex
\section{Related Work}
\label{sec:background}

\noindent \textbf{Audio Remixing.} Highlighting a mixed audio track is identical to rebalancing its individual sources, \ie, transferring from one mixing style to another. In previous research, music mixing~\cite{koo2023music,vanka2024diff,martinez2020deep,ramirez2019modeling} has been extensively studied, including creative manipulations that shape a song's emotive and sonic identity. Reproducing the mixing style of a target song typically involves balancing tracks using audio effects to achieve harmony and aesthetic appeal, often through knowledge-based~\cite{parker2022physical} or learning-based~\cite{steinmetz2022style,martinez2020deep,vanka2024diff} approaches. Related but different from these methods that focus on music, we handle a broad range of sounds, including speech, music, and sound effects. Moreover, we propose to use visual cues in videos to guide the highlighting process. %

\vspace{0.05in}

\noindent \textbf{Video Highlight and Saliency Detection.} Web videos, often created by professionals, are typically edited, such as trimming to capture key moments or adjusting camera focus to highlight engaging regions. This has led to the development of video understanding tasks, including video highlight detection~\cite{lei2021detecting,moon2023query,lin2023univtg,liu2022umt}, which identifies key temporal segments, and video saliency prediction~\cite{itti1998model,kroner2020contextual,harel2006graph,Jiang_2018_ECCV,jain2021vinet}, which identifies salient regions within a scene.
Developing methods to smartly emphasize the right visual content~\cite{si2023freeu,huang2025fresca} in a video and the development of corresponding benchmark datasets~\cite{lei2021detecting,gitman2014semiautomatic} have become essential areas of study.
In our work, we focus on the mirror side of the problem: assuming the video is already well-curated to visually convey highlight information, as in movies, we leverage this visual narrative to highlight the audio stream accordingly.

\vspace{0.05in}

\noindent\textbf{Audio-Visual Learning.}
Exploring connections between acoustic and visual signals has been widely studied across various tasks, including audio-visual localization~\cite{qian2020multiple,hu2020discriminative,tian2021cyclic,mo2022localizing,huang2023egocentric,tian2018audio,tian2020unified}, 
representation learning from cross-modal supervision~\cite{owens2018audio,gao2020visualechoes,afouras2020selfsupervised},
audio-visual learning for 3D scenes~\cite{liang2023neural,liang23avnerf,huang2024modeling}, audio-spatialization leveraging visual cues~\cite{gao20192,morgado2018self,garg2021geometry},
and sound generation from videos~\cite{chen2017deep,chen2024action2sound,owens2016visually,liang2024language}.
Differently, we tackle a new challenging audio-visual learning task, visually-guided acoustic highlighting.

Most closely related to our task is visually-guided audio separation~\cite{gao2018learning,owens2018audio,zhao2018sound,huang2024scaling,gao2021visualvoice,huang2023davis}, which addresses a specific case of our problem by isolating a target sound and suppressing others to zero, thereby achieving separation. However, these methods lack a well-defined output for a balanced mix and do not address the creation of poorly mixed inputs. Our approach offers a new perspective by focusing on remixing audio to produce a coherent, visually aligned output.

%% file: sec/3_method.tex
\section{Approach}
\label{sec:approach}

\begin{figure*}[t]
    \centering
    \includegraphics[width=1\linewidth]{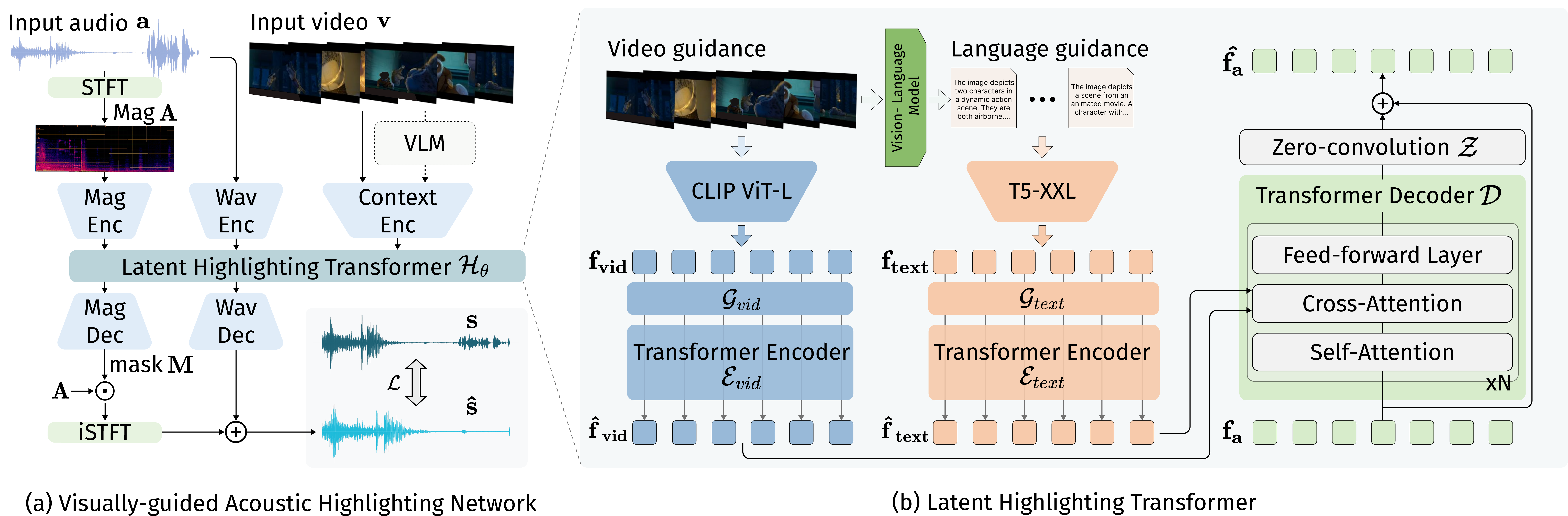}
    \caption{Overview of \ourmodel: (a) Our model takes a poorly mixed audio waveform as input and produces the highlighted audio using a dual U-Net architecture. For simplicity, skip connections are omitted in the illustration. (b) The latent highlighting transformer incorporates vision and text encoders to integrate temporal information, guiding the transformer decoder to transform audio features effectively.
    }
    \label{fig:method}
\end{figure*}

We introduce the task of visually-guided acoustic highlighting, which aims to transform audio using visual guidance to achieve an appropriate highlighting effect. In Sec.~\ref{subsec:task}, we elaborate on the task formulation and Sec.~\ref{sec:model} discusses design of our proposed multimodal model \ourmodel.

\subsection{Task Formulation}
\label{subsec:task}

Let $ \mathbf{v} \in  \mathbb{R}^{T_v \times H \times W \times 3}$ represent the visual stream, a sequence of $T_v$ RGB images, and let $\mathbf{a} \in  \mathbb{R}^{T_a}$ be the corresponding audio sequence. Here $\mathbf{a}$ is considered a poorly mixed audio sequence that lacks the intended highlighting effect, as is common in raw daily recordings where audio is directly captured without deliberate curation.
Our goal is to predict an audio signal $ \mathbf{s} $ that preserves the content of $ \mathbf{a}$ but conveys the appropriate highlighting effects. 
In other words, we learn a mapping from the poorly mixed audio and the video sequence to a corresponding highlighted audio signal:
\vspace{-0.2in}
\begin{align}
    \mathbf{a}, \mathbf{v} \mapsto \mathbf{s}.
    \label{eq:mapping}
    \vspace{-0.2in}
\end{align}

Creating a highlighting effect in audio involves rebalancing the sources at different levels to reflect their relative prominence. When describing system design next we will assume access to data tuples of the form $ ( \mathbf{a}, \mathbf{v}, \mathbf{s})$. 
How such a dataset is created and utilized in practice will be the subject of our discussion in Sec.~\ref{sec:dataset}.

\subsection{\ourmodel: Visually-guided Acoustic Highlighting}\label{sec:model}

To address the problem formulated in \cref{eq:mapping}, we focus on two primary design choices: (i) how to structure the audio framework to accept poorly mixed audio as input and produce highlighted audio, and (ii) how to effectively incorporate contextual information, such as video streams or other modalities like text, to guide the acoustic highlighting. In this section, we elaborate on our framework design, presenting a flexible approach that includes both an audio backbone and a context-aware module. The overall architecture of \ourmodel is illustrated in \cref{fig:method}.

\subsubsection{Audio Backbone}
The audio backbone is designed to produce an output with the same shape as the input, making U-Net~\cite{ronneberger2015u} architectures particularly suitable. We consider two common types of audio input: time-domain and frequency-domain. Frequency-domain input, often used in audio-visual separation tasks~\cite{zhao2018sound,tian2021cyclic,chen2023iquery}, captures distinct frequency patterns of sounds, while time-domain input, frequently employed in audio-only separation~\cite{luo2019conv,subakan2021attention}, provides higher accuracy in reconstructing the final waveform.
In this work, we unify the advantages of both input types within our audio backbone, offering flexibility for future studies to utilize either or both. Based on the HybridDemucs architecture~\cite{defossez2021hybrid}, we implement a dual U-Net model with two branches: one for spectrogram inputs and the other for waveform inputs.

\noindent \textbf{Spectrogram U-Net Encoder.}  Given an input audio signal $\mathbf{a} \in \mathbb{R}^{T_a}$, we apply a Short-Time Fourier Transform (STFT) to $\mathbf{a}$ with a window size of 4096 and hop length of 1024 to obtain its spectrogram. Specifically, we use the magnitude spectrogram as input, denoted by $\mathbf{A}$. In the original HybridDemucs~\cite{defossez2021hybrid}, the spectrogram is normalized using its mean and standard deviation. However, we omit this normalization, as we found that mean normalization can significantly suppress ambient sounds, reducing the model’s sensitivity to sound effects.
The magnitude encoder consists of 5 layers, with each layer reducing the number of frequency bins by a factor of 4, except for the final layer, which reduces it by a factor of 8. After passing through the magnitude encoder, the frequency dimension is reduced to 1, aligning it with the output shape from the waveform branch. 
Details of the encoder design can be found in the supplementary materials.

\noindent \textbf{Waveform U-Net Encoder.}  In this framework, the waveform branch acts as a residual path to capture fine-grained temporal details. To facilitate processing, we normalize the waveform input $\mathbf{a}$. The encoder design mirrors that of the spectrogram U-Net encoder, with the main difference being the use of 1D convolutions instead of 2D convolutions.

\noindent \textbf{Latent Highlighting Module.} With both the magnitude and waveform embeddings in the same shape, we add them element-wise to create a unified audio embedding. An additional encoder layer then reduces the temporal dimension by half, producing $\mathbf{f_a} \in \mathbb{R}^{C_a \times L}$, where $L$ represents the temporal dimension of the latent audio features. To transform $\mathbf{f_a}$  into highlighted audio representations, we design a latent highlighting module that incorporates both the audio features and contextual information $\mathbf{c}$ (such as the encoded features of video streams or other multi-modal input) to output the features representing the highlighted audio, denoted as: 
\vspace{-0.05in}
\begin{equation} 
\mathbf{\hat{f}_a} = \mathcal{H}_\theta(\mathbf{f_a}, \mathbf{c}),
\label{eq:latent}
\vspace{-0.05in}
\end{equation}
where $\theta$ is the model parameters. Since both the latent audio features and the contextual input are temporal signals, we utilize a transformer-based framework to process them effectively, as depicted in \cref{fig:method}(b).

\noindent \textbf{Waveform/Spectrogram U-Net Decoder.} The refined features $\mathbf{\hat{f}_a}$ will first pass through an additional decoder layer to double the temporal length. Next, the output features serve as the input to both the waveform and spectrogram U-Net decoders, each mirroring the structure of the corresponding encoder. The spectrogram decoder outputs a predicted ratio mask, denoted as $\mathbf{M}$, which represents the highlighting information. We multiply the mask with the original magnitude spectrogram element-wise to obtain a refined magnitude $\mathbf{M} \odot \mathbf{A}$. Then we apply inverse STFT on this refined magnitude, using the phase information from the input to reconstruct the output waveform.
In addition, the audio output from the waveform decoder is then combined with the spectrogram-based waveform output, producing the final prediction $\mathbf{\hat{s}}$.

\subsubsection{Latent Highlighting Transformer}
We now discuss the design of latent highlighting module $\mathcal{H}_\theta$ introduced in \cref{eq:latent}.

Latent audio features $\mathbf{f_a}$ from the audio backbone capture temporal and semantic characteristics of the poorly mixed audio. To transform these features into representations that convey appropriate highlighting effects, we consider two key insights:
(i) Audio captures information from the entire surrounding environment, while the visual field-of-view is narrower, focusing on salient regions and content. This necessitates leveraging the temporal dynamics of the visual context as guidance for acoustic highlighting.
(ii) In movies, complex interactions often occur between different sources (such as speech, music, and sound effects), with music saliency, in particular, driven by emotional cues. Relying solely on visual signals may not fully convey these nuanced relationships. This prompts the question: can additional modality enhance this process?
To study this, we design a transformer-based latent highlighting module $\mathcal{H}_\theta$ that can flexibly incorporate various types of temporal context, such as video streams or text captions.

\noindent \textbf{Context Encoding.} Given the video sequence $ \mathbf{v}$, we use CLIP ViT-L/14~\cite{radford2021learning} to transform each frame into a feature vector, denoted as $\mathbf{f}_{\text{vid}} \in \mathbb{R}^{C_{\text{vid}} \times T_v}$. To address the second insight mentioned above, we incorporate text captions as an additional modality. Vision Language Models (VLMs) have demonstrated impressive capabilities in summarizing images and reasoning about text. We leverage text captions as a bridge to convey deeper sentiment and context beyond raw visual features.
To generate captions automatically for each frame, we use InternVL2-8B~\cite{chen2024far}. Each caption is then embedded using T5-XXL encoder~\cite{raffel2020exploring}, resulting in textual embeddings denoted as $\mathbf{f}_{\text{text}} \in \mathbb{R}^{C_{\text{text}} \times T_v}$.
Since raw frame and text features are extracted at a per-frame level and lack temporal interaction, we apply a transformer encoder for each modality to capture the temporal context, denoted as $\mathcal{E}_{\text{vid}}$ and $\mathcal{E}_{\text{text}}$. To preserve temporal order, we add sinusoidal positional encoding~\cite{vaswani2017attention} to the input of each transformer encoder layer. 
We can concisely define contextual information encoding as a sequence of the following operations:
\vspace{-0.05in}
\begin{equation} 
\mathbf{\hat{f}}_i = \mathcal{E}_i(\mathcal{G}_i(\mathbf{f}_i)),
\vspace{-0.05in}
\end{equation}
where $i \in \{\text{video, text}\}$, $\mathcal{G}_i(\cdot)$ is a linear projection layer to project $C_i$ to $C$, the same channel dimension as $\mathbf{f_a}$.

\noindent \textbf{Context-aware Acoustic Highlighting.}  We use a transformer decoder, $\mathcal{D}$, to generate highlighted acoustic representations. Sinusoidal positional encoding is added to $\mathbf{f_a}$. The transformer decoder $\mathcal{D}$ consists of multiple layers, each containing a self-attention layer, a cross-attention layer, and a feed-forward layer. Rather than directly treating the decoder’s output as the final prediction, we interpret it as an offset to the original features, adding it back to $\mathbf{f_a}$ to preserve the audio’s semantic content while adjusting inter-source differences. Additionally, we incorporate a zero-initialized convolution layer~\cite{zhang2023adding}, denoted $\mathcal{Z}(\cdot)$. This layer is a 1×1 convolution with both weights and biases initialized to zero. The overall process is described as:
\vspace{-0.05in}
\begin{equation} 
\mathbf{\hat{f}_a} = \mathbf{f_a} + \mathcal{Z}(\mathcal{D}(\mathbf{f_a}, \mathbf{\hat{f}}_i)),
\vspace{-0.05in}
\end{equation}
where the context $\mathbf{\hat{f}}_i$ can include visual, textual, or both types of contextual information. In relation to \cref{eq:latent}, $\mathcal{H}_\theta$ acts as the integrative component that connects  $\mathcal{D}$, $\mathcal{E}_{\text{vid}}$, and potentially $\mathcal{E}_{\text{text}}$.
 
\subsubsection{Training and Inference} 
\label{subsec:training}
Our \ourmodel framework takes the input audio $\mathbf{a}$ along with visual context $\mathbf{v}$ or its textual captions to predict the highlighted audio $\mathbf{\hat{s}}$. The loss is computed at the waveform level and backpropagated through the network.
In this work, we use a multiscale STFT (MR-STFT) loss~\cite{yamamoto2021parallel} between the predicted audio $\mathbf{\hat{s}}$ and the ground truth audio $\mathbf{s}$, which is implemented by computing the $\ell_1$ distance between their amplitude spectrograms:
\vspace{-0.05in}
\begin{equation}
    \mathcal{L} = \text{MR-STFT}(\mathbf{\hat{s}},\mathbf{s}).
\vspace{-0.05in}
\end{equation}
The window sizes are set to 2048, 1024, and 512. It is worth noting that the training loss is intentionally simple, and any arbitrary waveform or spectrogram loss could be applied. We demonstrate that even a standard loss can effectively drive training, leaving further exploration of loss design to future work. 

At test-time, given badly mixed input audio and the associated video frames as context, \ourmodel outputs well-highlighted audio  that is coherent both temporally and semantically with the provided visual guidance. 

%% file: sec/4_dataset.tex
\begin{table*}[!ht]
    \caption{Main comparison: The best results are highlighted in \textbf{bold}, while the second best are highlighted with \underline{underline}. We report metrics on waveform distance, semantic alignment, and time alignment. All results are multiplied by 100. }
    \vspace{-0.05in}
    \centering
    \footnotesize
    \begin{tabularx}{0.9\textwidth}{lXccccc}
        \toprule
        Method & & MAG $\downarrow$ & ENV$\downarrow$ & KLD$\downarrow$ & $\Delta$IB$\downarrow$  & W-dis$\downarrow$\\
        \midrule
         \textit{Poorly Mixed Input} & & {$22.69$} & {$6.30$} & {$20.61$} & {$1.52$} & {$1.94$} \\
DnRv3~\cite{warcharasupat2024remastering}+CDX~\cite{Fabbro2023TheSD} & & {$26.32~(-16\%)$} & {$7.62~(-21\%)$}  & \underline{$15.87~(+23\%)$}  & \underline{$1.78~(-17\%)$}  & {$2.84~(-46\%)$}\\
        Learn2Remix~\cite{yang2022don} & & {$19.07~(+16\%)$} & \underline{$4.16~(+34\%)$} & {$61.76~(-199\%)$} & {$8.27~(-444\%)$} & \underline{$1.20~(+38\%)$}\\
        LCE-SepReformer~\cite{jiang2024listen} & & \underline{$17.18~(+24\%)$} & {$4.28~(+32\%)$} & {$30.99~(-50\%)$}  &{$1.88~(-24\%)$}  & {$1.28~(+34\%)$}  \\
        \ourmodel (Ours) & & {$\mathbf{10.08~(+56\%)}$}  & {$\mathbf{3.43~(+46\%)}$} & {$\mathbf{11.01~(+47\%)}$} & {$\mathbf{0.80~(+47\%)}$} & {$\mathbf{0.79~(+59\%)}$} \\
        \bottomrule
    \end{tabularx}
    \label{tab:main_comparison}
    \vspace{-0.05in}
\end{table*}

\section{\ourdataset}
\label{sec:dataset}

It is worth reiterating that training our model requires access to badly mixed input audio, well-highlighted output audio and the associated visual frames, as shown in \cref{eq:mapping}. Our key observation is that movies  serve as a reliable source of well-mixed data, implicitly conveying what good highlighting and its synchronization with video sounds like. 
Our final requirement of having access to the corresponding badly mixed input audio is satisfied through the data modification process described in this section. It essentially involves demixing, adjusting and then remixing movie audio so as to disturb its original highlighting effect.

\begin{figure}[!t]
    \centering
    \includegraphics[width=1\linewidth]{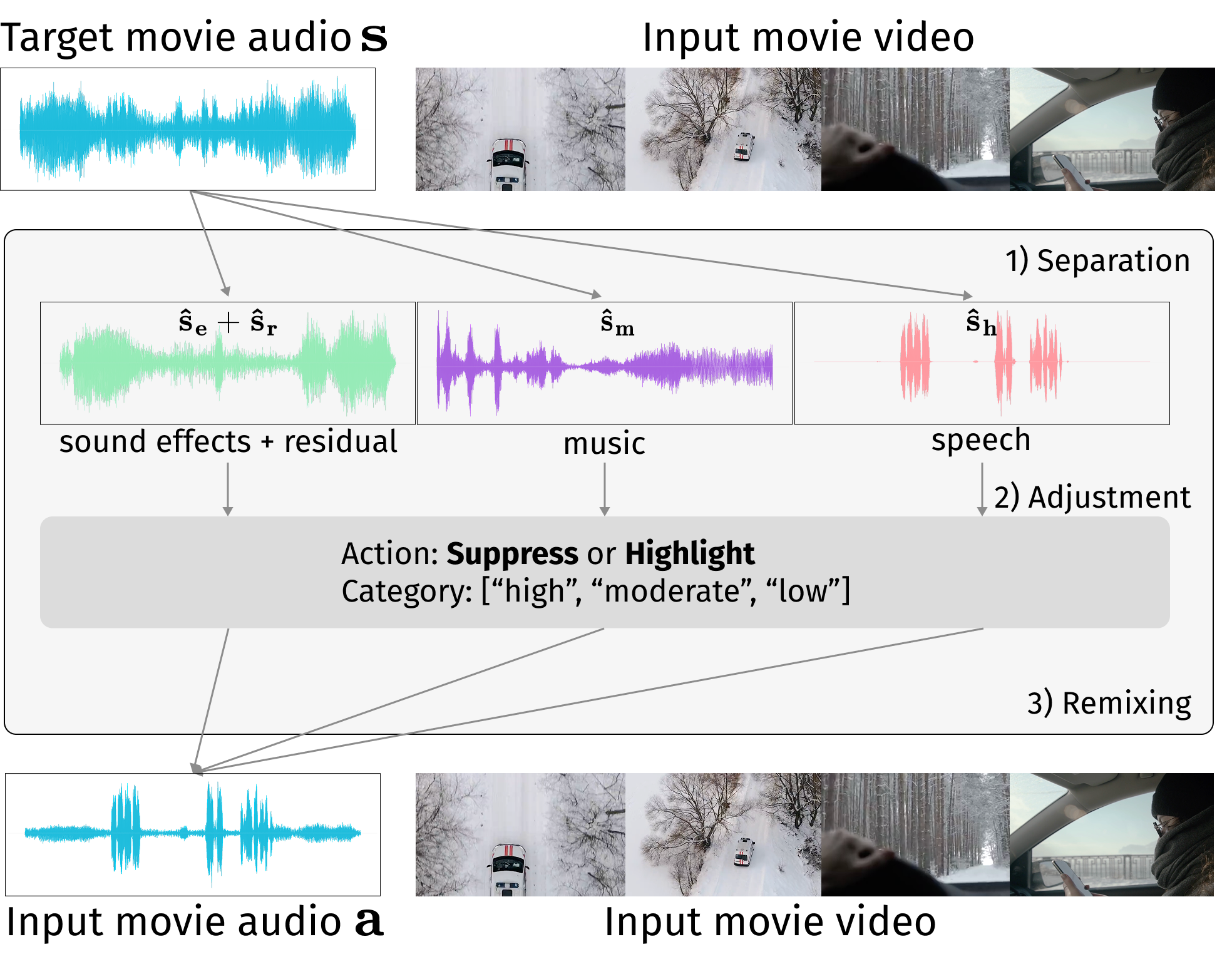}
    \vspace{-0.2in}
    \caption{We generate poorly mixed audio from the well-mixed movie audio through the following steps: 1) \textbf{Separation}: We separate the ground truth movie audio into individual tracks for speech, music, and sound effects, allowing for some imperfections in the separation process; 2) \textbf{Adjustment}: For each separated track, we apply either suppression or emphasis, with the intensity selected from three levels: [high, moderate, low]; 3) \textbf{Remixing}: Finally, we combine the adjusted tracks through simple addition to create the poorly mixed input audio. 
    }
    \label{fig:datacreation}
    \vspace{-0.2in}
\end{figure}

We select the CMD~\cite{bain2020condensed} as our data source, which includes 33,976 clips from 3,605 diverse movies spanning various genres%
, countries, and decades, covering salient parts of each film. Each clip is approximately two minutes long.  We concentrate on films tagged in the “Action” category, ensuring good presence of multiple acoustic sources beyond speech and music. As a movie in CMD may belong to multiple categories, our selection still remains diverse and covers a range of genres. We leave further expansion of the dataset to future work. To prepare the data for training and evaluation, we segment each movie clip into 10-second segments and extract the video stream at 1 fps using ffmpeg, while filtering out segments that lack an audio stream\footnote{All data collection and processing was done at the University of Rochester}.

\vspace{0.05in}

\noindent \textbf{Separate, Adjust, and Remix.} Given a high-quality movie audio $\mathbf{s}$, we prepare the poorly mixed input $\mathbf{a}$ through a three-step process as shown in \cref{fig:datacreation}:
\begin{enumerate}
\vspace{0.05in}
    \item \textbf{Separation.} In practice, audio may consist of an infinite variety of sources, making it impractical to separate and remix every possible source individually. We follow the Cinematic Sound Demixing Challenge~\cite{Fabbro2023TheSD}, which segments audio into three broad categories: \textit{speech}, \textit{music}, and \textit{sound effects}. Accordingly, we apply a three-stem separation model trained for cinematic audio source separation on the DnR v3 dataset~\cite{warcharasupat2024remastering}, to decompose $\mathbf{s}$ into three substreams: $\mathbf{\hat{s}_h}$ (speech), $\mathbf{\hat{s}_m}$ (music), and $\mathbf{\hat{s}_e}$ (sound effects). Additionally, we calculate any residual component, $\mathbf{\hat{s}_r}$, to ensure that $\mathbf{s} = \mathbf{\hat{s}_h} + \mathbf{\hat{s}_m} + \mathbf{\hat{s}_e} + \mathbf{\hat{s}_r}$. This formulation guarantees that even if the separation is imperfect, the sum of all components matches the original audio track.

\vspace{0.05in}

    \item  \textbf{Adjustment.} Using these imperfect separations, we alter their original relative levels, creating an input audio signal that intentionally mismatches the video’s highlighting effect. Specifically, we adjust the relative loudness of each stream. We first measure the original loudness of each separated source using the \textit{pyloudnorm} library~\cite{steinmetz2021pyloudnorm}. For the source with the highest loudness, we apply a \textbf{“Suppress”} action, reducing its loudness by a randomly selected strength from the categories [high, moderate, low]. For the other two sources, we apply a \textbf{“Highlight”} action, increasing their loudness by a value chosen from [high, moderate, low].
    To retain the original mixture’s content, we use the combined track $\mathbf{\hat{s}_e} + \mathbf{\hat{s}_r}$ for the sound effects input signal. We implement the loudness adjustments as follows: [high, moderate, low] for highlighting corresponds to increases of $\{12, 9, 6\}$ dB, while for suppressing, we apply decreases of $\{-12, -9, -6\}$ dB.
\vspace{0.05in}

    \item \textbf{Remixing.} After adjusting loudness, we remix the three sources linearly to create a poorly mixed input that contrasts with the ground truth highlighting effect.
\end{enumerate}

\vspace{0.05in}

Following this procedure, we generate input audio for each video clip, resulting in 15,078/1,927/1,789 clips for train/validation/test sets, respectively.

%% file: sec/5_experiments.tex
\section{Experiments}
\label{sec:experiments}

\subsection{Experimental Setting}
\label{subsec:setting}
\noindent\textbf{Implementation Details.}  In our experimental setup, the audio waveform is sampled at 44 kHz in stereo. We convert the input to mono by averaging the two stereo channels. Within the encoders, we set the dimensionality of the audio latent representation $\mathbf{f_a}$ to $C_a = 768$, with the original channel dimensions for visual and text features set to $C_{\text{vid}} = 768$ and $C_{\text{text}} = 4096$, respectively. During training, we use a batch size of 12 per GPU and the Adam optimizer with a learning rate of 0.0001. The model is trained for 200 epochs. All experiments are conducted on two RTX 4090 GPUs, with training taking approximately 18 hours to complete.

\vspace{0.05in}

\noindent \textbf{Evaluation Metrics.} 
We employ the following groups of objective metrics to evaluate output quality:
\begin{itemize}
    \item \textbf{Waveform distance}: The simplest way to assess the closeness of the prediction to the target is through waveform distance. We use magnitude distance (MAG)~\cite{xu2021visually} to evaluate audio quality in the time-frequency domain and envelope distance (ENV)~\cite{liang23avnerf} to assess quality in the time domain.
    \item \textbf{Semantic alignment}: Since our goal is to adjust the relative distribution of audio across three categories: human speech, music, and sound effects. We apply KL divergence (KLD)~\cite{liu2023audioldm,vyas2023audiobox} using the pre-trained PaSST~\cite{koutini22passt} model to compare the label distributions of the target and generated audio. Additionally, given that the video provides guidance, we assess audio-to-video semantic relevance using the ImageBind~\cite{girdhar2023imagebind} model, calculated as the cosine similarity between audio and video embeddings, denoted as IB score. Since we have the target movie audio, we use the difference between the target and predicted IB scores:
    \begin{equation}
        \Delta \text{IB} =  \text{IB}(\mathbf{v},\mathbf{s}) - \text{IB}(\mathbf{v},\mathbf{\hat{s}}).
    \end{equation}
    \item \textbf{Time alignment}: The relative variation between underlying sources (speech, music, and sound effects) can lead to significant timing differences, as each track follows its own temporal pattern. To test how well the model highlights all sources, we measure the minimum cost to align the predicted audio distribution with the target distribution. This is quantified using Wasserstein Distance (W-dis)\footnote{\url{https://en.wikipedia.org/wiki/Wasserstein_metric}.}.
\end{itemize}

\begin{table}[!t]
    \caption{Ablation study on different context types. We compare a no-context baseline with models using semantic (single frame or text caption) and temporal context (multiple frames or captions).}
     \vspace{-0.05in}
    \centering
    \footnotesize
    \begin{tabularx}{0.4\textwidth}{lXccc}
        \toprule
        Context & & MAG $\downarrow$ & KLD$\downarrow$ & $\Delta$IB$\downarrow$\\
        \midrule
        No Context & & $10.35$ & $11.95$ & $0.99$ \\
        \midrule
        \quad +Semantic Vision & & $10.35$ & $11.67$ & $0.91$ \\
        \quad +Semantic Text & & $10.32$ & $11.83$ & $0.84$ \\
        \midrule
        \quad +Temporal Vision & & $10.24$ & $11.18$ & $0.88$ \\
        \quad +Temporal Text & &$10.08$ & $11.01$ & $0.80$ \\
        \bottomrule
    \end{tabularx}
    \label{tab:conditions}
\end{table}

\subsection{Baselines}
This is a novel task with no prior works. We adapt several methods for relevant and fair comparison with \ourmodel:
\begin{itemize}
    \item \noindent \textbf{Poorly Mixed Input}: This is the manually created poorly mixed input according to our dataset creation strategy, serving as a reference point for comparison.
    \item \noindent \textbf{DnRv3~\cite{warcharasupat2024remastering}+CDX~\cite{Fabbro2023TheSD}}: To remix speech, music, and sound effects from the input and generate highlighted audio, we include an empirical baseline that adheres to the loudness distribution of these sources as specified by the CDX~\cite{Fabbro2023TheSD} challenge. We first apply the DnRv3~\cite{warcharasupat2024remastering} separator to split the input audio into three tracks: speech, music, and sound effects. Next, we sample loudness values for each track according to their respective distributions. Finally, we adjust the loudness of each source and remix them to create the output audio.
    \item \noindent \textbf{Learn2Remix~\cite{yang2022don}}:  Learn2Remix (L2R) utilizes ConvTasNet~\cite{luo2019conv} as its backbone model to predict and remix different audio sources within feature spaces, making it well-suited for our task of adjusting and rebalancing the underlying speech, music, and sound effects. In our implementation, we adopt the more advanced SepReformer~\cite{shin2024separate} model to replace the ConvTasNet backbone. We use the official code and train it on our dataset.
    \item \noindent \textbf{Listen, Chat and Edit (LCE)~\cite{jiang2024listen}}: LCE is a text-guided sound mixture editor capable of performing various audio editing tasks, such as adjusting the volume of specific sources based on text instructions. However, in our setup, we assume that explicit instructions on which sounds to highlight are unavailable and instead should be inferred from visual cues. To ensure a fair comparison, we provide text captions as guidance for LCE. Originally, LCE uses ConvTasNet~\cite{luo2019conv} and Sepformer~\cite{subakan2021attention} as the SoundEditor models. For a fair comparison, we also replace the backbone with the SepReformer~\cite{shin2024separate} model.
\end{itemize}

\begin{table}[!t]
    \caption{Ablation study on transformer encoders $\mathcal{E}_{\text{vid}}$ and $\mathcal{E}_{\text{text}}$. V and T represent vision frames and text captions, respectively. Note that the text captions are obtained automatically from the video.}
     \vspace{-0.05in}
    \centering
    \footnotesize
    \begin{tabularx}{0.47\textwidth}{lccXccc}
        \toprule
        \#layers & \#params & context & & MAG $\downarrow$ & KLD$\downarrow$ & W-dis$\downarrow$ \\
        \midrule
        0  & 55.3M & V & & $10.36$ & $10.91$ & $0.83$ \\
        3 & 61.6M & V & & $10.24$ & $11.18$ & $0.81$ \\
        6 & 67.9M & V & & $10.69$ & $12.42$ & $0.83$\\
        \midrule
        0  & 55.4M & T & & $10.66$ & $12.53$ & $0.85$ \\
        3 & 61.7M & T & & $10.34$ & $11.75$ & $0.81$ \\
        6 & 68.0M & T & & $10.08$ & $11.01$ & $0.79$ \\
        \bottomrule
    \end{tabularx}
    \label{tab:layers}
     \vspace{-0.1in}
\end{table}

\begin{figure*}[!t]
    \centering
    \includegraphics[width=0.85\linewidth]{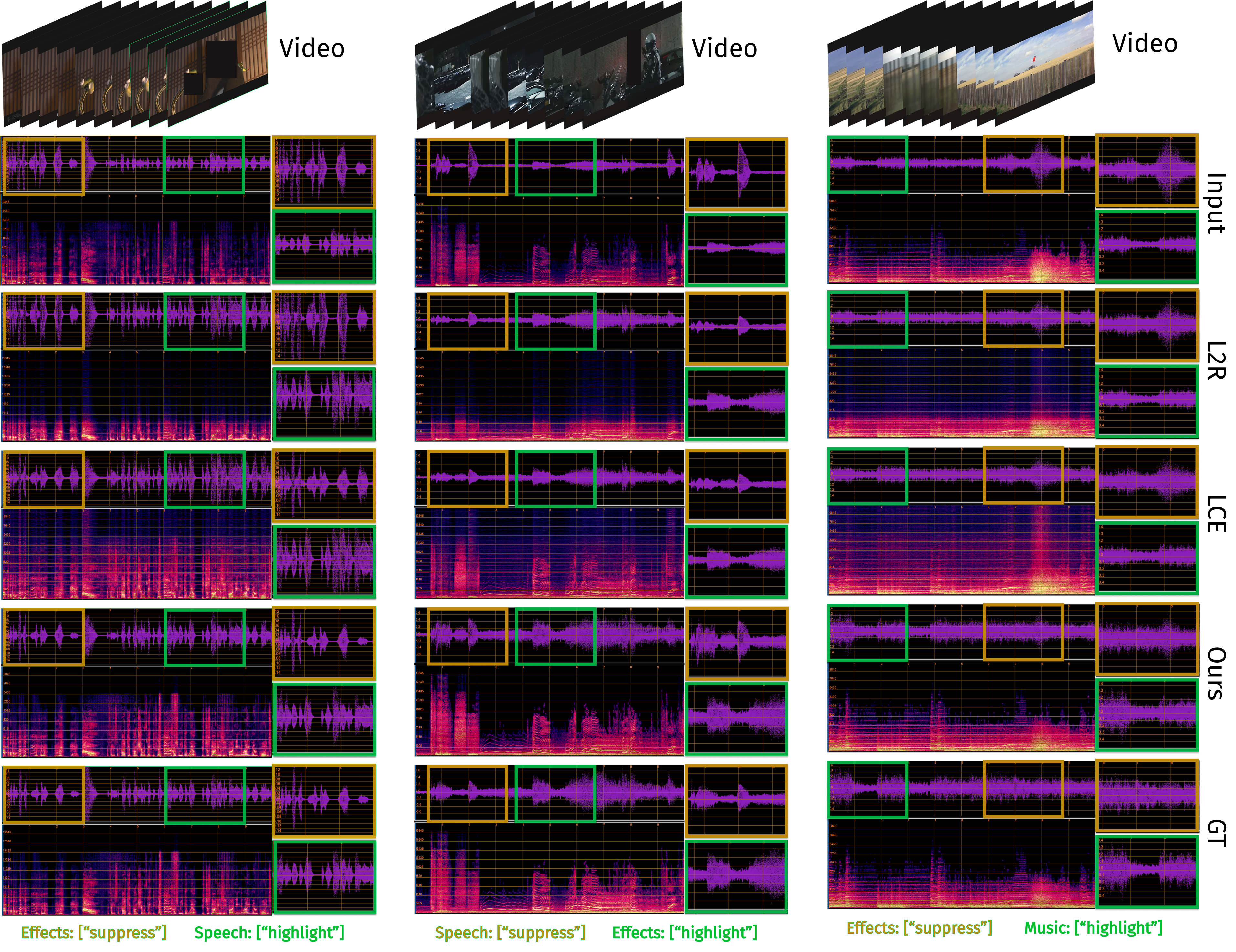}
     \vspace{-0.05in}
    \caption{We perform a qualitative comparison by visualizing the waveform and magnitude spectrograms of the highlighted audio results from different methods, along with the input and ground truth. Our method produces results that are closest to the movie GT. The \textcolor{orange}{orange box} denotes suppressed snippets, and \textcolor{olive}{green box} indicates highlighted snippets. }
    \label{fig:qualitative}
     \vspace{-0.2in}
\end{figure*}

\subsection{Quantitative Results}

\subsubsection{Comparison with Baselines}
\noindent We compare our method with the baselines, and the results are presented in \cref{tab:main_comparison}. The empirical baseline DnRv3+CDX performs worse than the input in waveform distance and time alignment metrics because it relies on a non-specific statistical distribution rather than data-dependent remixing. However, it outperforms the other two baselines in KLD, which we hypothesize is due to the versatile loudness distributions of the movies in CDX; remixing the speech, music, and sound effects at those levels shifts the distribution toward real movies.
On the other hand, Learn2Remix, an audio-only baseline, struggles to enhance audio without guidance. It improves the poorly mixed input in terms of waveform distance, which occurs because it learns the global loudness distribution of our dataset. However, it fails to achieve the necessary semantic alignment, reinforcing the need for contextual input. 
For LCE, we utilize text captions as guidance for acoustic highlighting, resulting in better outcomes compared to Learn2Remix. However, it still underperforms significantly when compared to our approach. This is because it is not a method designed for acoustic highlighting and lacks the ability to capture the global trends required for the task. In contrast, \ourmodel demonstrates strong performance across all metrics, showcasing the effectiveness of our proposed framework.

\subsubsection{Ablations}
\label{subsubsec: ablations}
In this section, we review the design of context encoding and context choice, providing further insights into the task setup.

\noindent \textbf{Does Contextual Information Matter?} In \cref{tab:conditions}, we present a naive baseline that does not utilize any contextual guidance. This means that the model relies solely on the input audio to learn how to highlight relevant information. However, when we incorporate a single frame or its corresponding text caption, the semantic alignment metrics, KLD and $\Delta$IB, show improvement. This indicates that context plays a crucial role in enhancing the model's performance. Since audio is a time sequence, the highlighting effects should ideally capture certain temporal patterns. We further conduct ablation experiments using the full length of video frames and captions, referring to this as temporal vision or text. The performance shows a significant boost, underscoring the importance of temporal context.

\noindent \textbf{Number of Transformer Encoder Layers.} We evaluate the impact of encoding temporal context by varying the number of transformer encoder layers. Without transformer encoders, video and text features are encoded frame by frame, which results in a lack of interaction across time steps. As shown in Table \cref{tab:layers}, we find that increasing the number of layers generally improves performance, suggesting that temporal context reasoning is essential for effectively understanding video-level content. Specifically, we observe continuous improvement when using text context, while performance with vision context initially improves but then deteriorates when the number of layers further increases, which we hypothesize that the CLIP vision features are already compact.

\begin{table}[!t]
    \caption{Ablation study on dataset difficulty. We report the performance of Input (-I) and our Predictions (-P) at the three different levels of dataset difficulty.}
     \vspace{-0.05in}
    \centering
    \footnotesize
    \begin{tabularx}{0.47\textwidth}{lXccccc}
        \toprule
        Level & & MAG $\downarrow$ & ENV$\downarrow$ & KLD$\downarrow$ & $\Delta$IB$\downarrow$  & W-dis$\downarrow$\\
        \midrule
        High-I & & $27.70$ & $7.64$ & $32.52$ &$2.35$ & $2.38$ \\
        High-P & & $12.03$ & $3.93$ & $16.11$ & $1.25$ & $0.97$ \\
        \midrule
        Moderate-I & & $22.70$ & $6.25$ & $20.59$ & $1.50$ & $1.92$\\
        Moderate-P & & $8.73$ & $3.23$ & $11.08$ & $0.81$ & $0.65$\\
        \midrule
        Low-I & & $16.40$ & $4.48$ & $9.89$ & $0.75$ & $1.37$\\
        Low-P & & $9.55$ & $3.20$ & $7.16$ & $0.35$ & $0.80$\\
        \bottomrule
    \end{tabularx}
    \label{tab:difficulty}
     \vspace{-0.1in}
\end{table}

\noindent \textbf{Analysis of Dataset Difficulty.} In \cref{sec:dataset}, we outline three levels of adjustments that can be made during the creation of the dataset, which are randomly selected. In \cref{tab:difficulty}, we provide an ablation of the impact of these difficulty levels. We create three test sets that consist solely of low, moderate, or high levels of adjustments. Our observations show continuous improvements in metrics as the difficulty of the dataset decreases. This supports both the design of our dataset and metrics, as well as the generalization capability of training on randomly selected levels.

\begin{figure}[!t]
    \centering
    \includegraphics[width=.85\linewidth]{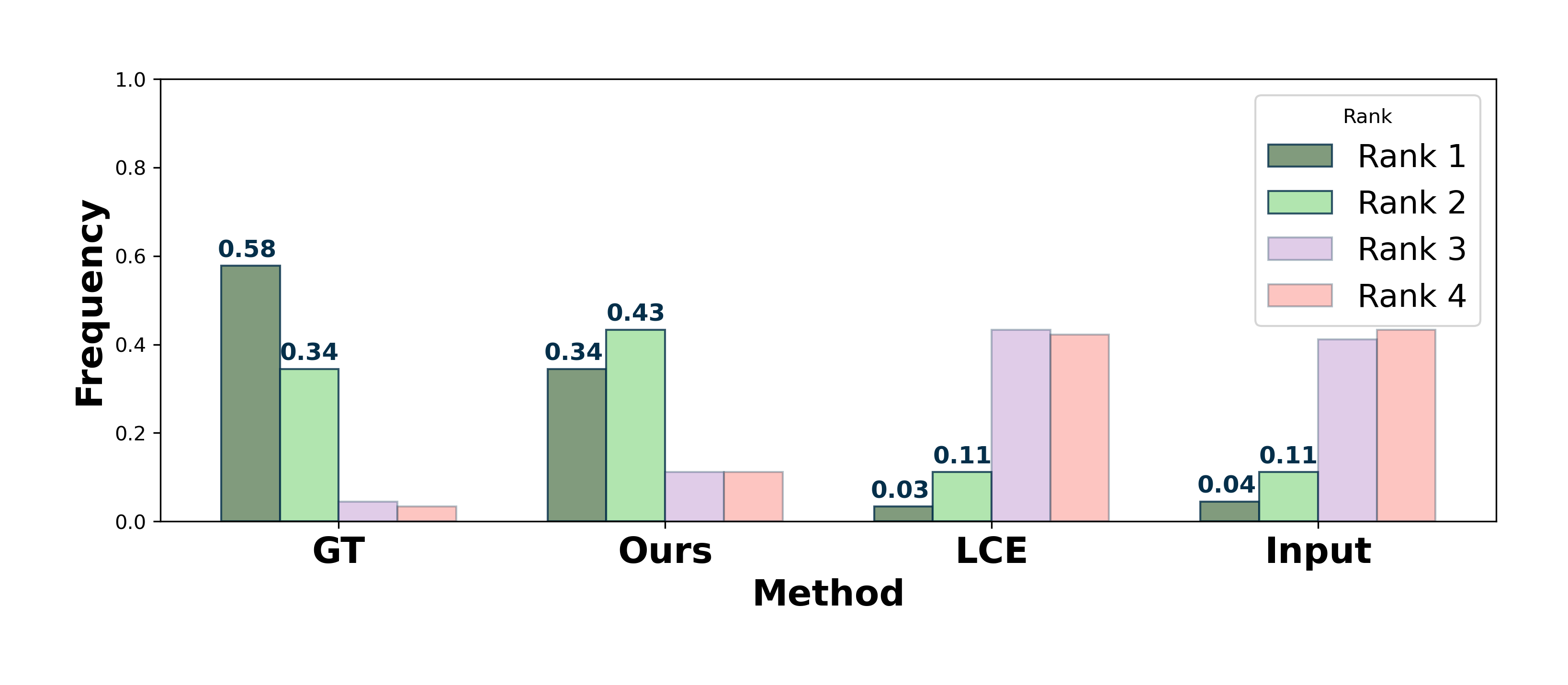}
    \vspace{-0.2in}
    \caption{Subjective test: We ask users to rank the four methods based on audio-visual balance to evaluate acoustic highlighting.}
    \label{fig:subjective}
     \vspace{-0.15in}
\end{figure}

\subsection{Qualitative Analysis}
\label{subsec:qualitative}

\noindent \textbf{Qualitative Visualizations.} We display the magnitude spectrograms and waveforms of the highlighted audio produced by various methods, along with the input and ground truth in \cref{fig:qualitative}. These visualizations illustrate that our method effectively captures temporal variations and performs acoustic highlighting across speech, music, and sound effect sources.

\vspace{0.05in}
    
\noindent \textbf{Subjective Test.}
We conduct a subjective test to compare the highlighting results of our model with those of the LCE baseline, as well as the input and ground truth. Nine participants evaluated ten videos, each featuring four different audio tracks generated by various methods. They ranked the four methods based on perception of the balance between audio and visual quality. Our method achieves a top-2 ranking rate of 77\%, outperforming the LCE baseline and the input by 63\% and 62\%, respectively, as shown in \cref{fig:subjective}. Interestingly, our method even surpasses the GT for 34\% of the videos, indicating strong highlighting performance, comparable to actual films at times.

\begin{figure}[t]
    \centering
    \includegraphics[width=1\linewidth]{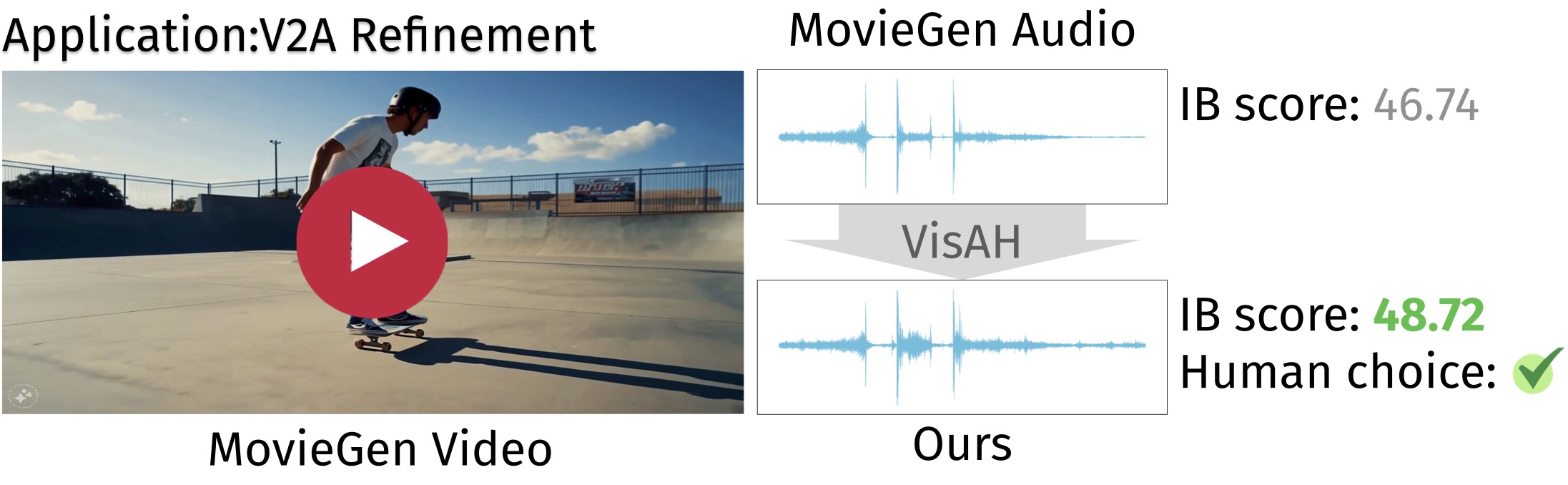}
    \vspace{-0.25in}
    \caption{We demonstrate that our VisAH method enhances the quality of video-to-audio generation results.}
    \label{fig:applicaiton}
    \vspace{-0.15in}
\end{figure}

\noindent \textbf{Application: Refinement of Video-to-Audio Generation.} Our VisAH has several potential downstream applications, one of which is refining video-to-audio generation. In \cref{fig:applicaiton}, we demonstrate that by using the audio from MovieGen~\cite{polyak2024movie} as input, along with the video as guidance, our VisAH produces audio that achieves a better IB score. This indicates enhanced audio-visual alignment, and human preferences confirm these improvements.  We encourage the readers to see and listen to examples on our demo webpage in the attached supplementary materials.

%% file: sec/6_conclusion.tex
\section{Conclusion}
\label{sec:conclusion}

We presented a new task---visually-guided acoustic highlighting---to bridge the gap between visual and acoustic saliency in video content. To address this task, we have proposed \ourmodel, a transformer-based multimodal framework that uses visual information to guide audio highlighting. By leveraging movies for free supervision, we develop a pseudo-data generation process that simulates real-world video quality, allowing for a labor-free training setup.
Our evaluations show that our approach outperforms several baselines in both objective and human perceptual assessments. This framework enhances the alignment of audio-visual cues, offering a more cohesive viewing experience.

%% file: sec/X_suppl.tex
\clearpage
\setcounter{page}{1}
\setcounter{figure}{6}
\maketitlesupplementary

\section{Project Page}
\label{sec:demo}
We have created a project page (\url{https://wikichao.github.io/VisAH/}) to illustrate our method and showcase our results. \textbf{We strongly encourage readers to visit this webpage and use headphones}. Please note that the webpage may not be fully compatible with the Safari browser; therefore, we recommend using Google Chrome for an optimal viewing experience. On the demo page, we show the following applications:

\begin{itemize}
    \item \textbf{Comparisons to Other Methods.} We present examples from \ourdataset, showcasing the following: the input poorly mixed video (which is created through the process described in Sec.4, the highlighting results produced by LCE~\cite{jiang2024listen}, the outputs from our VisAH model, and the original movie clips for comparison.

    \item \textbf{Video-to-Audio (V2A) Generation Refinement.} Generating audio from video has recently gained popularity due to impressive video generation results and the growing demand for an immersive audio-visual experience. Existing V2A models, such as Seeing-and-Hearing~\cite{xing24seeing} and the more recent MovieGen~\cite{polyak2024movie}, have demonstrated promising outcomes. However, these methods primarily focus on generating temporally aligned audio for videos, which can sometimes neglect the subtle differences between audio sources. Our approach, inspired by cinematic techniques, serves as a post-processing method to enhance audio quality in these cases.

    \item \textbf{Real Web Video Refinement.} Unlike movies, web videos are often recorded in less controlled environments, which can lead to undesirable effects. For example, viewers may experience an overpowering personal voice in egocentric videos or focus on distracting sound sources due to distance or background noise. In this context, we apply our model to web videos, aiming to deliver an improved cinematic-like audio-visual experience.
\end{itemize}

\begin{figure}[ht]
    \centering
    \includegraphics[width=1\linewidth]{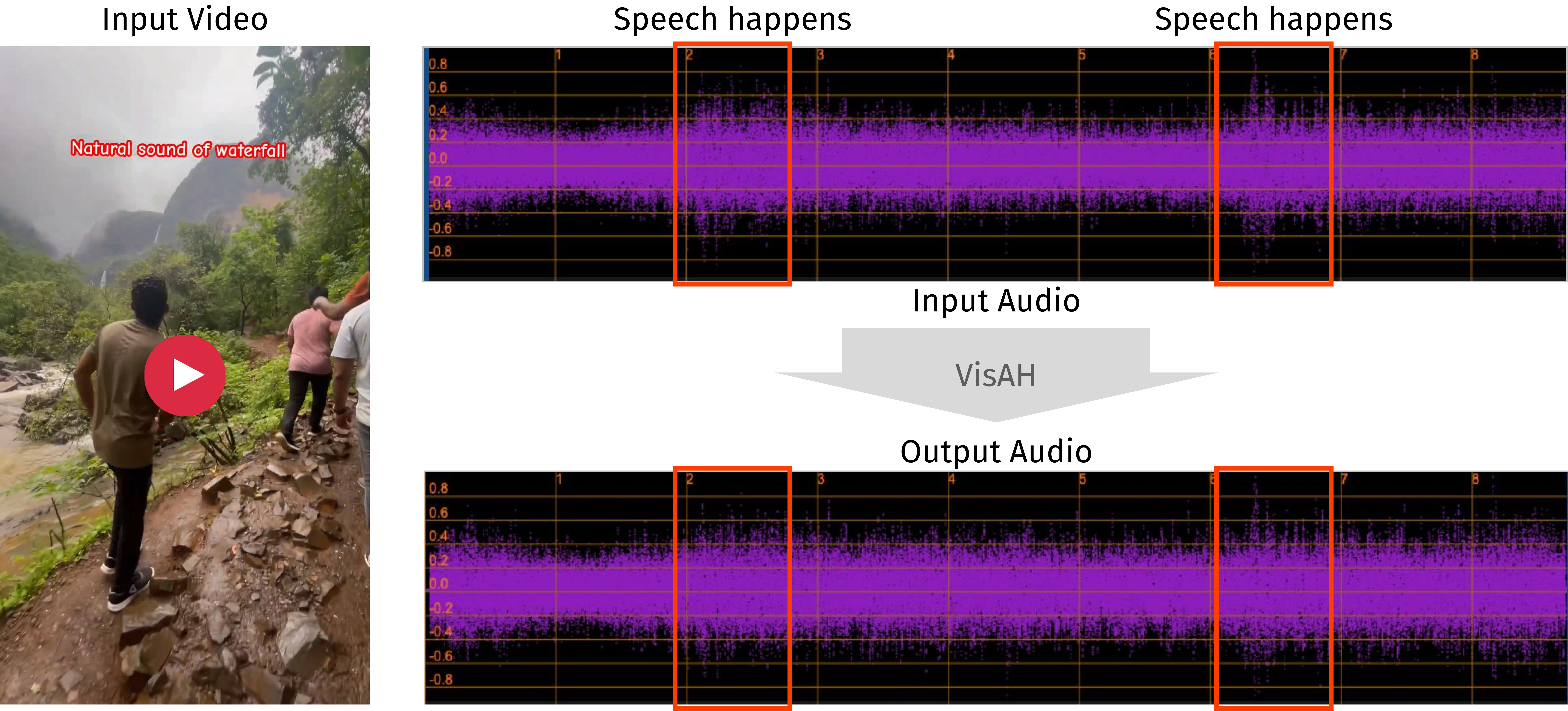}
    \caption{Failure case analysis: the sound effect (waterfall) overwhelms the speech.}
    \label{fig:failure_audio}
\end{figure}

\begin{figure}[!h]
    \centering
    \includegraphics[width=1\linewidth]{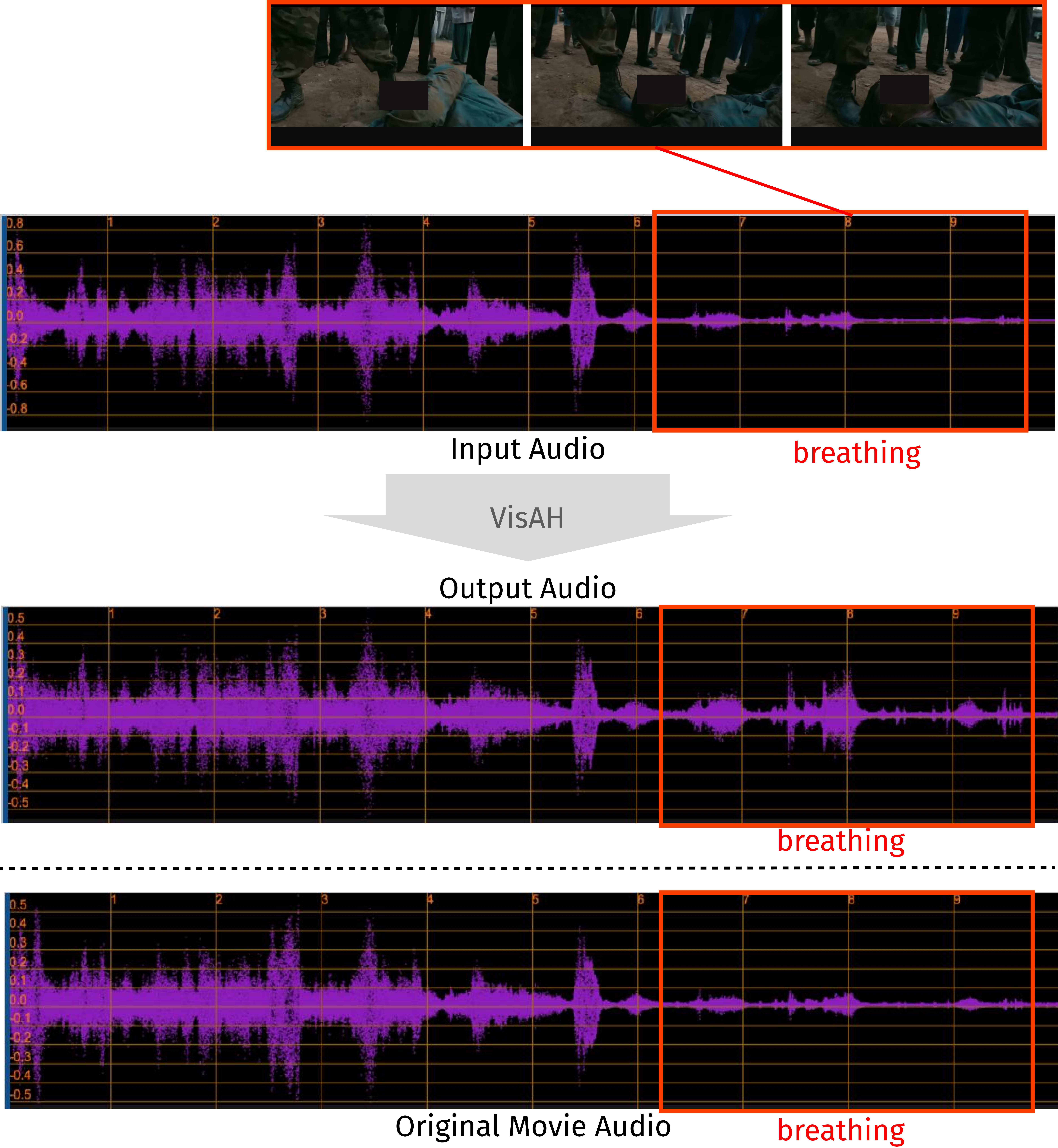}
    \caption{Failure cases analysis: Our method highlights the breathing sound based on the video context but diverges from the movie audio ground truth.}
    \label{fig:failure_video}
\end{figure}

\section{Failure Case Analysis}
\label{sec:failure}
While our \ourmodel model is effective at highlighting audio guided by video content, there are scenarios where it might fail. Here, we provide case studies to illustrate the conditions under which such failures occur.

In \cref{fig:failure_audio}, the video captures a natural waterfall scene with people hiking. The audio stream predominantly features the sound of the waterfall, with occasional moments of speech. Ideally, our \ourmodel model should balance these two audio sources to enhance the audio-visual experience. However, due to the overwhelming dominance of the waterfall sound, the speech becomes difficult to perceive. This results in the model failing to properly highlight the speech. As shown in \cref{fig:failure_audio}, the input and output audio remain similar in this case, highlighting the challenge of separating and emphasizing speech under such conditions.

In \cref{fig:failure_video}, we present an example where our method fails to align perfectly with the original movie ground truth. Specifically, the breathing sound between 7 and 10 seconds is not emphasized in the movie’s ground truth audio. However, the corresponding video frames during this period show close-up shots of a man's face, visually depicting the breathing action. Given these video conditions, our method predicts output audio that highlights the breathing sound, aligning with the visual context but diverging from the original movie audio. This failure highlights the need for a deeper understanding of movie content to achieve better alignment with the intended audio design.

\section{Subjective Test Design}
\label{sec:subjective_platform}

\begin{figure}[t]
    \centering
    \includegraphics[width=1\linewidth]{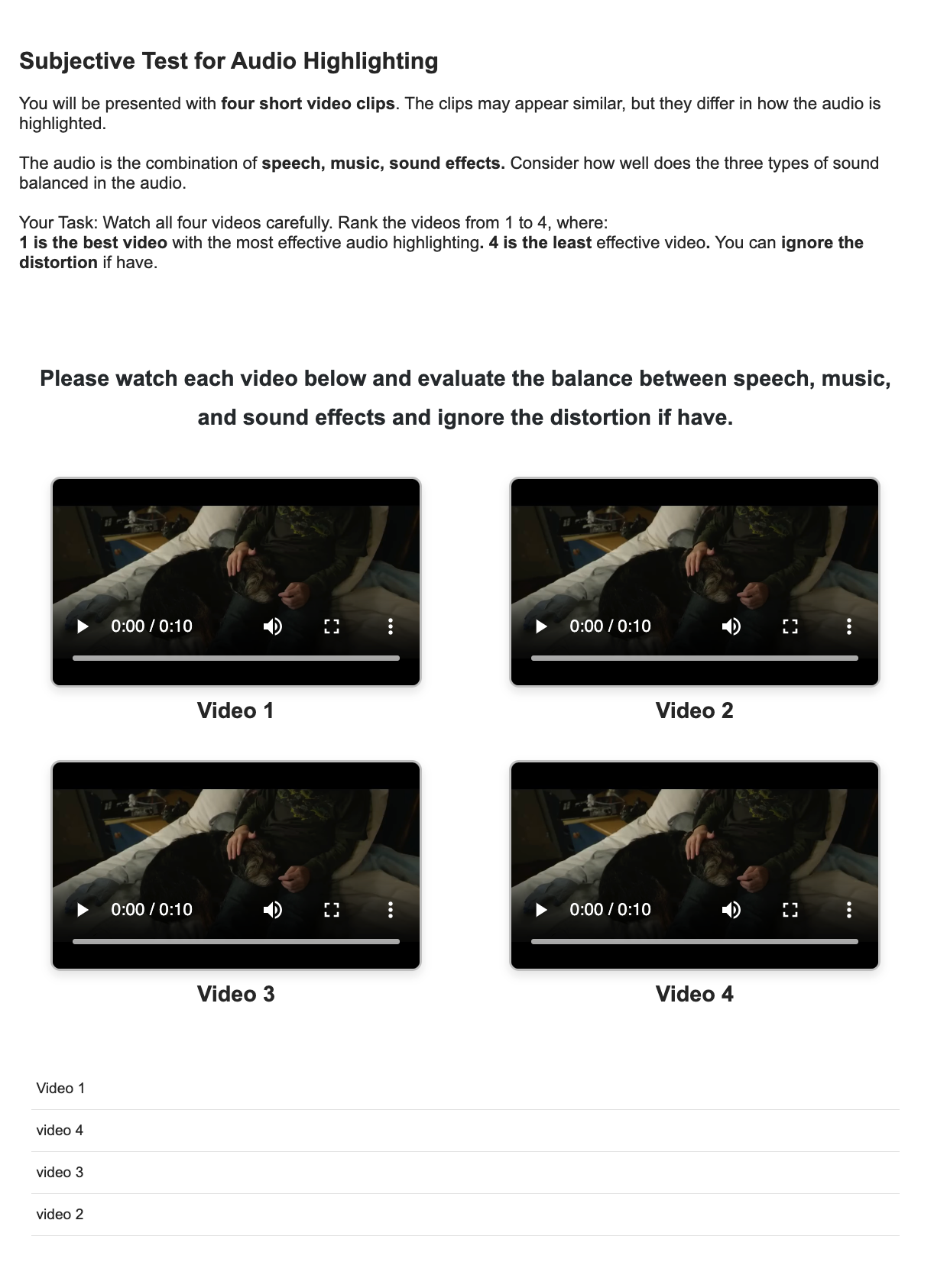}
    \caption{Screenshot of subjective test interface.}
    \label{fig:interface}
\end{figure}

We illustrate the interface design of our subjective test in \cref{fig:interface}. The instructions emphasize that users should evaluate whether the speech, music, and sound effects in the videos are well-balanced and acoustically pleasing, and whether the audio aligns effectively with the video content.

Participants are shown four videos: the poorly mixed input, the best-performing baseline (LCE), our method, and the movie ground truth. After watching all the videos, users are asked to rank them from 1 to 4, with 1 being the most effective in audio highlighting and 4 being the least effective. The analysis of the ranking results is presented in Fig.~5.

\section{Network Details}
\label{sec:network}
\begin{figure}[t]
    \centering
    \includegraphics[width=1\linewidth]{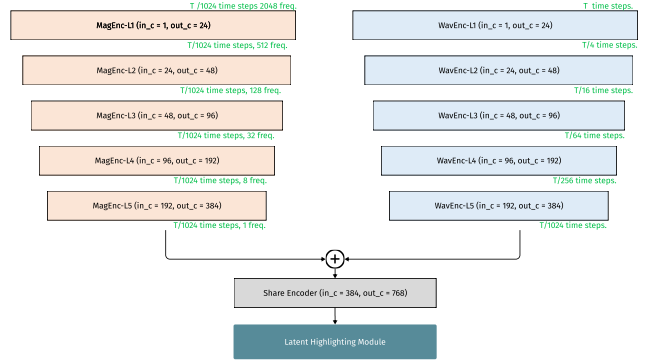}
    \caption{Design of magnitude and waveform encoders. Each encoder consists of five layers. The features from the waveform and magnitude encoders are combined through element-wise addition after the fifth layer, followed by an additional layer to encode the fused features.}
    \label{fig:encoder}
\end{figure}

We detail the design of the magnitude and waveform encoders, along with their input and output dimensions. As illustrated in \cref{fig:encoder}, each encoder consists of five layers, and the output shapes for both branches after the fifth layer are identical. At each layer, the output features are used for skip connections (not shown in the figure). This design facilitates straightforward element-wise addition of the two branches. The fused feature is then processed through a shared encoder layer before being passed to the latent highlighting module.
Similarly, the magnitude and waveform decoders mirror the architectures of the encoders in reverse order.

\section{Loss Function Details}
\label{sec:loss}
Here, we give a more detailed illustration on the MR-STFT (Multi-Resolution Short-Time Fourier Transform) loss function used for training the model. The MR-STFT loss is implemented by computing the $\ell_1$ distance between the amplitude spectrograms of the predicted signal \(\hat{s}\) and the ground truth signal \(s\). Mathematically, the loss function can be expressed as:
\[
L_{\text{MR-STFT}}(\hat{s}, s) = \sum_{k=1}^{K} \left\| |STFT_k(\hat{s})| - |STFT_k(s)| \right\|_1,
\]
where \(STFT_k(\cdot)\) denotes the Short-Time Fourier Transform with the \(k\)-th window size, and \(|\cdot|\) represents the magnitude of the spectrogram. The window sizes are set to 2048, 1024, and 512, corresponding to different resolutions of the spectrogram. This multi-resolution approach allows the loss function to capture both fine-grained and coarse-grained spectral details of the signals. It is worth noting that the training loss is intentionally simple, and any arbitrary waveform or spectrogram loss could be applied. We demonstrate that even a standard loss, such as the MR-STFT loss, can effectively drive training and lead to high-quality results.

\begin{figure}[h]
    \centering
    \includegraphics[width=1\linewidth]{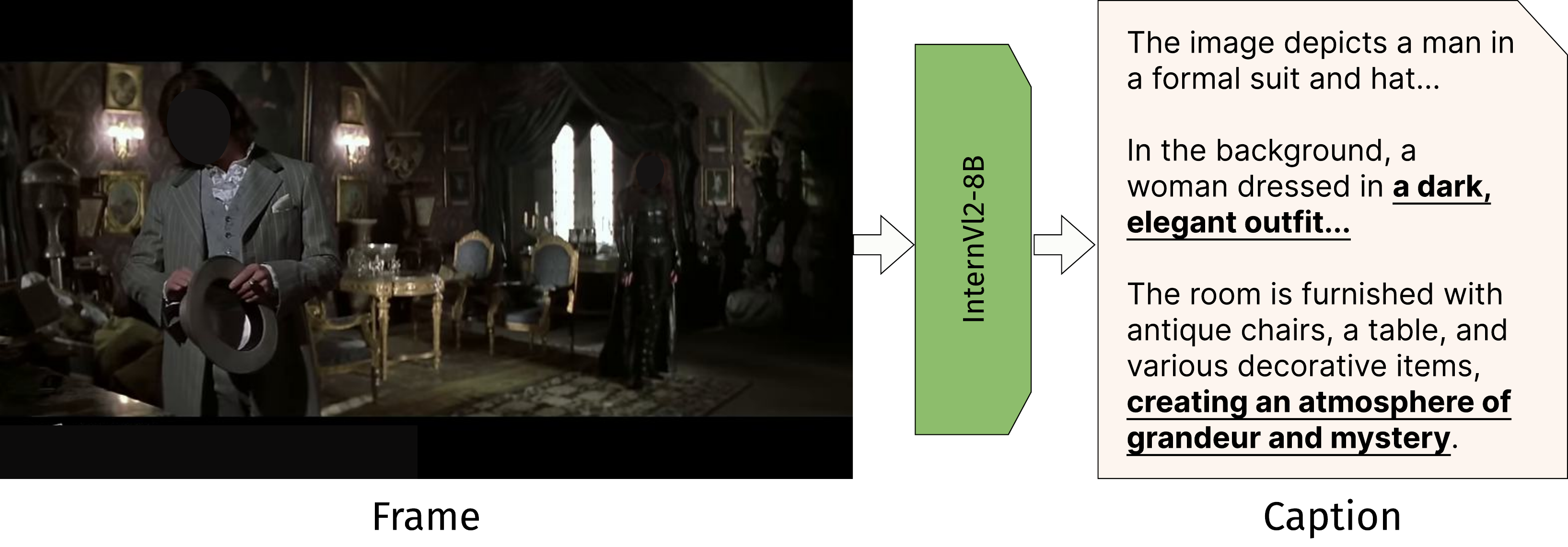}
    \caption{An example of a video frame and its generated caption.}
    \label{fig:caption}
\end{figure}

\section{Motivation for Text Condition.} Textual captions supplement video frames by leveraging strong reasoning capabilities of MLLMs. In \cref{fig:caption}, the caption generated by InternVL2-8B captures not only visual content, such as the appearance of individuals and room decorations, but also the scene's atmosphere, demonstrating the added semantic richness that textual information can provide. Moreover, it provides information more \textit{explicitly} (e.g. ``a dark, elegant outfit") than the visual encoder may extract. This supports the observation in \cref{tab:conditions} of why text conditioning outperforms visual signals. Regarding the performance metrics of the visual encoder in \cref{tab:layers}, we hypothesize that CLIP vision features are more compact, and the 1fps video sampling rate drops motion information. Consequently, vision features are easier to overfit, as observed with the peak performance when the number of vision encoder layers is 3, and more encoder layers cause smoothing. 
To address this, we can try adopting a higher framerate (\eg, 8fps) or exploring motion-aware architectures such as temporal transformers or 3D convolutions, which better model temporal dynamics while minimizing computational overhead. Learned downsampling can be another potential solution.

\begin{figure}[t]
    \centering
    \includegraphics[width=1\linewidth]{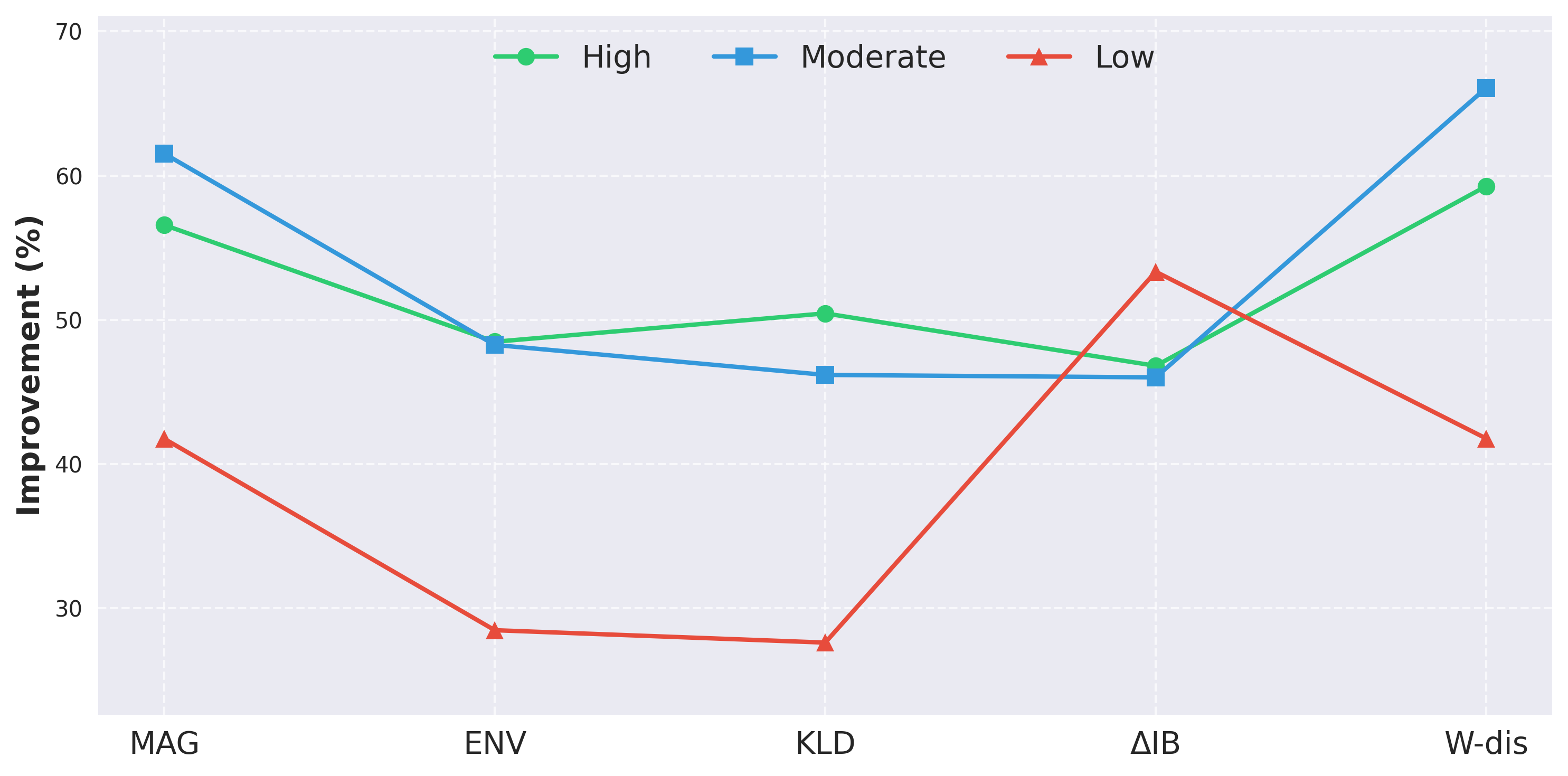}
    \caption{The improvement trend across the three difficulty levels is evaluated over five metrics.}
    \label{fig:trend}
\end{figure}

\section{Inference Time Comparison} 
The inference times for VisAH, LCE, and L2R audio backbone are 0.028s, 0.017s, and 0.018s, respectively. While our method requires more time, it remains efficient for practical applications.

\section{Analysis of Dataset Difficulty}
\label{sec:difficulty}
We visualize the improvement trends in \cref{fig:trend} across different levels of dataset difficulty, as discussed in Sec~5.3.2 and shown in Tab.~4. The magnitude of improvement is similar for the high and moderate difficulty levels, demonstrating that our method is robust in highlighting audio sources, even when they are highly suppressed. In contrast, the lower improvement observed for the low-difficulty level is attributed to the fact that the input audio is already relatively close to the ground truth and thus inherently conveys the ground truth highlighting effects to some extent. Consequently, the potential for improvement is reduced in this group.

\section{Limitations and Future Works}
\label{sec:future works}
Our method leverages versatile temporal conditions as guidance for audio highlighting, outperforming baseline methods and demonstrating applicability to real-world scenarios, including transferring knowledge from movies to daily and generated videos. However, there are areas where improvements can be made:

\noindent \textbf{(i) Multimodal Condition Fusion.} In our approach, we use either the video or its corresponding frame captions as guidance, achieving effective highlighting results. However, integrating these two modalities remains an open challenge. Text captions can infer the sentiment of the movie, complementing the video stream. Designing a more sophisticated strategy to fuse these modalities could enhance performance and remains an interesting direction for future research.

\noindent \textbf{(ii) Dataset Generation Strategy.} This paper introduces a three-step process for generating pseudo data through separation, adjustment, and remixing. While effective, each step can be further improved. For instance, employing multiple separators with varying granularity levels could offer greater flexibility and control. Additionally, replacing discrete loudness categories with continuous sampling could introduce more variability and challenge the model. Temporal loudness adjustments, such as varying the loudness at one-second intervals within a 10-second audio clip, could further enrich the dataset and present more complex training scenarios.

In summary, this work presents a novel task---visually guided acoustic highlighting---along with a versatile dataset generation process and a universal network. While our method demonstrates strong potential, many avenues for improvement remain, paving the way for future advancements in this area.